\documentclass[twoside]{article}
\usepackage{aistats2012arxiv}
\usepackage{amsmath,amsfonts,graphicx,natbib}
\usepackage{array}

\newcommand{\X}{\mathbf{X}}
\newcommand{\Y}{\mathbf{Y}}
\newcommand{\f}{\mathbf{f}}
\newcommand{\F}{\mathbf{F}}
\newcommand{\w}{\mathbf{w}}
\newcommand{\W}{\mathbf{W}}
\newcommand{\Z}{\mathbf{Z}}
\newcommand{\R}{\mathbf{R}}
\newcommand{\N}{\mathcal{N}}
\newcommand{\rb}{\boldsymbol{r}}
\newcommand{\el}{\langle}
\newcommand{\er}{\rangle}

\newcommand{\z}{\mathbf{z}}
\newcommand{\E}{\mathbf{E}}

\begin{document}

\twocolumn[

\aistatstitle{Bayesian Group Factor Analysis}

\aistatsauthor{Seppo Virtanen$^1$, Arto Klami$^1$, Suleiman A. Khan$^1$, Samuel Kaski$^{1,2}$}

\aistatsaddress{$^1$Aalto University School of Science\\Department of Information and Computer Science\\Helsinki Institute for Information Technology HIIT\\$^2$University of Helsinki}
]

\newcommand{\fix}{\marginpar{FIX}}
\newcommand{\new}{\marginpar{NEW}}

\begin{abstract}
  We introduce a factor analysis model that summarizes the
  dependencies between observed variable \emph{groups}, instead of
  dependencies between individual variables as standard factor
  analysis does. A group may correspond to one view of the same set of
  objects, one of many data sets tied by co-occurrence, or a set of
  alternative variables collected from statistics tables to measure
  one property of interest. We show that by assuming group-wise sparse
  factors, active in a subset of the sets, the variation can be
  decomposed into factors explaining relationships between the sets
  and factors explaining away set-specific variation. We formulate the
  assumptions in a Bayesian model which provides the factors, and
  apply the model to two data analysis tasks, in neuroimaging and
  chemical systems biology.
\end{abstract}

\section{Introduction}

Factor analysis (FA) is one of the cornerstones of classical data
analysis.  It explains a multivariate data set $\X \in \mathbb R^{N
  \times D}$ in terms of $K < D$ factors that capture joint
variability or dependencies between the $N$ observed samples
of dimensionality $D$. Each factor or component has a weight for each dimension,
and joint variation of different dimensions can be studied by
inspecting these
weights, collected in the loading matrix $\W \in \mathbb{R}^{D \times
  K}$. Interpretation is made easier in the sparse variants of FA
\citep{Knowles11,Paisley09,Rai09fa} which favor solutions with only a
few non-zero weights for each factor.

We introduce a novel extension of factor analysis, coined \emph{group
  factor analysis} (GFA), for finding factors
that capture joint variability between \emph{data sets} instead of
individual variables. Given a collection ${\X_1,...,\X_M}$ of $M$ data
sets of dimensionalities $D_1,...,D_M$, the task is to find $K <
\sum_{m=1}^M D_m$ factors that describe the collection and in
particular the dependencies between the data sets or views $\X_m$. Now every
factor should provide weights over the data sets, preferably again in
a sparse manner, to enable analyzing the factors in the same way as in
traditional FA.

The challenge in moving from FA to GFA is in how to make the factors
focus on dependencies between the data sets.
For regular FA it is sufficient to include a separate variance
parameter for each dimension. Since the variation independent of
all other dimensions can be modeled as noise, the factors will then
model only the dependencies. For GFA that would not be sufficient,
since the variation specific to a multi-variate data set can be more
complex. To enforce the factors to model only dependencies, GFA
hence needs to explicitly model the independent variation, or
structured noise, within each data
set. We use linear factors or components for that as well, effectively
using a principal component analyzer (PCA) as a noise model within
each data set.

The solution to the GFA problem is described as a set of $K$ factors
that each contain a projection vector for each of the data sets having
a non-zero weight for that factor. A fully non-sparse solution would
hence have $K \times M$ projection vectors or, equivalently, $K$
projection vectors over the $\sum_{m=1}^M D_m$-dimensional
concatenation of the data sources. That would, in fact, correspond to
regular FA of the feature-wise concatenated data sources.
They key in learning the GFA solution is then in correctly fixing the
sparsity structure, so that some of the components will start
modeling the variation specific to individual data sets while
some focus on different kinds of dependencies.

An efficient way of solving the Bayesian GFA problem can be
constructed by extending (sparse) Bayesian Canonical Correlation
Analysis \citep{Archambeau08} from two to multiple sets and by
replacing variable-wise sparsity by
group-wise sparsity as was recently done by
\citet{Virtanen11icml}. The model builds on insights from these
Bayesian CCA models and recent non-Bayesian group sparsity works
\citep{Jenatton10,Jia10}. The resulting model will operate on the
concatenation $\Y=[\X_1,...,\X_M]$ of the data sets, where the groups
correspond to the data sets. Then the factors in the GFA (projection
vectors over the dimensions of $\Y$) become sparse in the sense that
the elements corresponding to some subset of the data sets become zero,
separately for each factor. 
The critical question is to which extent is the model able to extract
the correct factors amongst the exponentially many alternatives that
are active in any given subset of data sets. We empirically demonstrate
that our Bayesian model for group-wise sparse factor analysis
finds the true factors even from fairly large number of data sets.

The main advantages of the model are that (i) it is conceptually very simple,
essentially a regular Bayesian FA model with group-wise sparsity, and
(ii) it enables tackling completely new kinds of data analysis
problems. In this paper we apply the model to two real-world example
scenarios specifically requiring the GFA model, demonstrating how the
GFA solutions can be interpreted. The model is additionally applicable to
various other tasks, such as learning of the subspace of
multi-view data predictive of some of the views, a problem addressed
by \citet{Chen10}.

The first application is analysis of fMRI
measurements of brain activity. Encouraged by the recent success in discovering
latent brain activity components in complex data setups
\citep{Lashkari10,Morup10,Varoquaux10},
we study a novel kind of an analysis setup where the same subject has
been exposed to several different representations of the same musical
piece. The brain activity measurements done under these different
conditions are considered as different views, and GFA reveals
brain activity patterns shared by subsets of different conditions. For
example, the model reveals ``speech'' activity shared by conditions
where the subject listened to a recitation of the song lyrics instead of
an actual musical performance.

In the second application drug responses are studied by modeling four
data sets, three of which contain gene expression measurements of
responses of different cell lines (proxies for three diseases; \citep{Lamb06}) and
one contains chemical descriptors of the drugs. Joint analysis of
these four data sets gives a handle on which drug descriptors are
predictive of responses in a specific disease, for instance.

\section{Problem formulation}

The group factor analysis problem, introduced in this work, is as
follows: Given a collection ${\X_1 \in \mathbb R^{N \times D_1},...,\X_M \in
  \mathbb R^{N \times D_M}}$ of data sets (or views) with $N$ co-occurring observations,
find a set of $K$ factors that describe the joint data set
$\Y=[\X_1,...,\X_M]$. Each factor is a sparse binary vector $\f_k \in
\mathbb R^{M}$ over the data sets, and the non-zero elements indicate
that the factor describes dependency between the corresponding
views. Furthermore, each active pair (factor $k$, data set $m$) is
associated with a weight vector $\w_{m,k} \in \mathbb R^{D_m}$ that
describes how that dependency is manifested in the corresponding data set $m$.
The $\w_{m,k}$ correspond to the factor loadings of regular FA, which
are now multivariate;
their norm reveals the strength of the factor and the vector
itself gives more detailed picture on the nature of the dependency.

The $\f$'s have been introduced to make the problem formulation and
the interpretations simpler; in the specific model we introduce next
they will not be represented explicitly. Instead, the weight vectors
$\w_{m,k}$ are instantiated for all possible factor-view pairs
and collected into a single loading matrix $\W$, which is then made
group-wise sparse.

\begin{figure*}[t]
\centering
\includegraphics[scale=1]{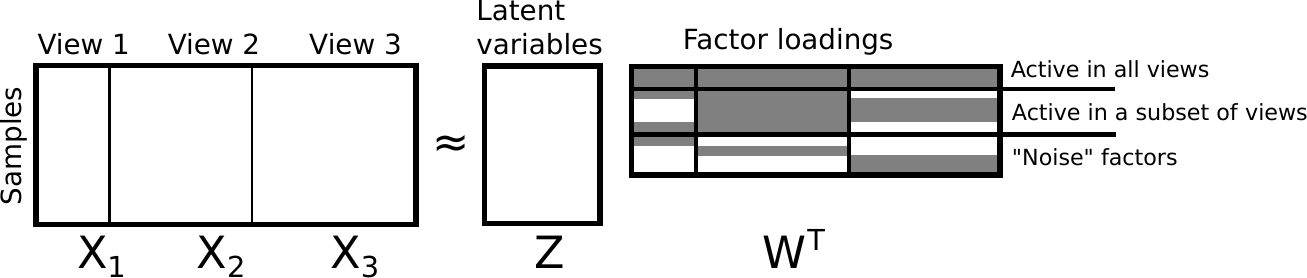}
\caption{Illustration of the group factor analysis of three data
  sets or views. The feature-wise concatenation of the data sets $\X_i$ is
  factorized as a product of the latent variables $\Z$ and factor
  loadings $\W$. The factor loadings are group-wise sparse, so that
  each factor is active (gray shading, indicating $\f_{m,k}=1$) only in some subset of views
  (or all of them). The factors active in just one of the views model
  the structured noise, variation independent of all other views, whereas the rest
model the dependencies. The nature
  of each of the factors is learned automatically, without needing to
  specify the numbers of different factor types (whose number would be
  exponential in the number of views) beforehand.}
\label{fig:factorization}
\end{figure*}

\section{Model}

We solve the GFA problem with a group-wise sparse matrix factorization
of $\Y$, illustrated in Figure~\ref{fig:factorization}. The variable groups correspond to the views $1,...,M$ and the
factorization is given by
\[
\Y \approx \Z \W^T,
\]
where we have assumed zero-mean data for simplicity. The factorization
gives a group-wise sparse weight matrix $\W \in \mathbb R^{D \times
  K}$ and the latent components $\Z \in \mathbb R^{N \times K}$. The
weight matrix $\W$ collects the factor- and view-specific
projection vectors $\w_{m,k}$ for all pairs for which $\f_{m,k}=1$.
The rest of the elements in $\W$ are filled with zeroes.

We solve the problem in the Bayesian framework, providing a generative
model that extracts the correct factors by modeling explicitly the
structured noise on top of the factors explaining dependencies. We assume the
observation model
\[
\Y = \Z\W^T + \E,
\]
where each row of the Gaussian noise $\E$ has diagonal noise covariance with
the diagonal of $[\sigma_1^2,...,\sigma_M^2]$ where $\sigma_m^2$ has been
repeated $D_m$ times. That is, every dimension within the same view
has the same residual variance, but the views may have different
variances. To complete the generative formulation, we assume the rows of $\Z$ to
have a zero-mean Gaussian distribution with unit covariance. That is,
each factor is represented with a Gaussian latent variable and all
factors are {\it a priori} independent.

The weight matrix $\W$ is made sparse by assuming a group-wise
automatic relevance determination (ARD) prior,
\begin{align*}
\alpha_{m,k} & \sim \text{Gamma}(a_0,b_0) \\
p(\W) & = \prod_{k=1}^{K}
\prod_{m=1}^M \prod_{d=1}^{D_m} \N(\w_{m,k}(d)|0,\alpha_{m,k}^{-1}) \; ,
\end{align*}
where $\w_{m,k}(d)$ denotes the $d$th element in the projection vector
$\w_{m,k}$, the vector corresponding to the $m$th view and $k$th
factor. The inverse variance of each vector is governed by the
parameter $\alpha_{m,k}$ which has a Gamma prior with a
small $a_0$ and $b_0$ (we used $10^{-14}$). The ARD makes groups
of variables inactive for specific factors by driving their $\alpha_{m,k}^{-1}$
to zero. The components used as modeling the structured noise within
each data set are automatically produced as factors active in
only one view.

Since the model is formulated through a sparsity prior we do not explicitly need to
represent $\F$ in the model.
It can, however, be created based on the factor-specific relative contributions
to the total variance of each view, obtained by
integrating out both $\z$ and $\W$. We set $\f_{m,k}=1$ if
\begin{equation}
\alpha_{m,k}^{-1} > \epsilon \left (\text{Tr}(\boldsymbol{\Sigma}_m) - \sigma_m^2 \right ) / D_m,
\label{eq:threshold}
\end{equation}
where $\text{Tr}(\boldsymbol{\Sigma}_m)$ is the total variance of the
$m$th view and $\epsilon$ is a small threshold constant.

The inference is based on a variational approximation.
We build on the approximation provided for Bayesian FA by
\citet{Luttinen10}, re-utilizing the EM-style update rules for all of
the parameters of the model.  The only differences are that the
posterior approximation for $\alpha$ needs to be updated for each
factor-view pair separately, $\sigma_m^2$ are view-specific instead
of feature-specific, and the parts of $\W$ corresponding to different
views are updated one at a time; the detailed update formulas are not
repeated here due to the close similarity.

To solve the difficult problem of fixing the rotation in factor
analysis models, we borrow ideas from the recent solution by
\citet{Virtanen11icml} for CCA models. Between each round of the
EM updates we maximize the variational lower bound with respect
to a linear transformation $\R$ of the latent subspace, which is
applied to both $\W$ and $\Z$ so that the product $\Z\W^T$ remains
unchanged. That is, $\hat \Z = \Z \R^T$ and $\hat \W^T = \R^{-T}
\W^T$. Given the fixed likelihood, the optimal $\R$ corresponds to a
solution best matching the prior that assumes independent latent
components, hence resulting in a posterior with maximally uncorrelated
components. We optimize for $\R$ by maximizing 
\begin{align}
L = &-\frac{1}{2} \text{Tr}( \R^{-1}\el \Z^T\Z \er \R^{-T} ) + C \log |\R| \notag \\
 &- \sum_{m=1}^M \frac{D_m}{2} \log \prod_{k=1}^{K} \rb_k^T \el \W_m^T\W_m \er \rb_k
\label{eq:rotation}
\end{align}
with the L-BFGS algorithm for unconstrained optimization.
Here $C= \sum_mD_m - N$, and $\rb_k$ is the $k$th column of $\R$, and the
$\el \Z^T\Z \er = \sum_n \el \z_n \z_n^T \er $ collects the second
moments of the factorization. Similar notation is used for $\W_m$, which indicates
the part of $\W$ corresponding to view $m$.

\subsection{Special cases and related problems}

When $D_m=1$ for all $m$ the problem reduces to regular factor
analysis. Then the $\w_{m,k}$ are scalars and can be incorporated into
$\f_k$ to reveal the factor loadings.

When $M=1$, the problem reduces to probabilistic principal component analysis
(PCA), since all the factors are active in the same view and they 
need to describe all of the variation in the single-view data set with linear
components.

When $M=2$, the problem becomes canonical correlation analysis (CCA) as
formulated by \citet{Archambeau08} and \citet{Virtanen11icml}. This is because
then there are only three types of factors. Factors active in both data
sets correspond to the canonical components, whereas factors active
in only one of the data sets describe the residual variation in each
view.  Note that most multi-set extensions of CCA applicable for $M>2$
data sets, such as those by \citet{Archambeau08,Deleus11}, do not solve the GFA
problem. This is because they do not consider components that could be
active in subsets of size $I$ where $2 \le I < M$, but instead
restrict every component to be shared by all data sets or to be specific
to one of them.

A related problem formulation has been studied in statistics under the
name multi-block data analysis. The goal there is to analyze connections
between blocks of variables, but again the solutions typically assume
factors shared by all blocks \citep{Hanafi06}. The
model recently proposed by \citet{Tenenhaus11} can find factors shared
by only a subset of blocks by studying correlations between block
pairs, but the approach requires specifying the subsets in advance.

Recently \citet{Jia10} proposed a multi-view learning model
that seeks components shared by any subset of views, by searching for a
sparse matrix factorization with convex optimization.  However, they
did not attempt to interpret the factors and only considered
applications with at most three views.

\citet{Knowles07} suggested that regular sparse FA \citep{Knowles11} could be
useful in a GFA-type setting. They applied sparse FA to analyzing
biological data with five tissue types concatenated in one feature
representation. In GFA analysis the tissues would be considered as
different views, revealing automatically the sparse factor-view
associations that can only be obtained from sparse FA after a separate
post-processing stage.  In the next section we show that directly
solving the GFA problem outperforms the choice of thresholding sparse
FA results.

\section{Technical demonstration}
\label{sec:toy}

For technical validation of the model, we applied it to simulated data
that has all types of factors: Factors specific to just one view,
factors shared by a small subset of views, and factors common to
most views. We show that the proposed model can correctly discover the
structure already with limited data, while demonstrating that possible
alternative methods that could be adapted to the scenario do not find
the correct result. We sampled $M=10$ data sets with dimensionalities
$D_m$ ranging between 5 and 10 ($\sum_m D_m=72$), using a manually
constructed set of $K=24$ factors of various types.

For comparing our Bayesian GFA model with alternative methods that
could potentially find the same structure, we consider the following
constructs:
\begin{itemize}
\item FA: Regular factor analysis for the concatenated data $\Y$. The
  model assumes the correct number of factors, $K=24$.
\item BFA: FA with an ARD prior for columns of $\W$, resulting in a Bayesian FA
  model that infers the number of factors automatically but assumes
  each factor to be shared by all views.
\item NSFA: Fully sparse factor analysis for $\Y$. We use the nonparametric sparse FA method by \citet{Knowles11}
  which has an Indian buffet process formulation for inferring the
  number of factors.
\end{itemize}
With the exception of the simple FA model, the alternatives are
comparable in the sense that they attempt to automatically infer
the number of factors, which is a necessary prerequisite for
modeling collections of several datasets, and that they are based
on Bayesian inference.

The solution for the GFA problem is correct if the model (i) discovers
the correct sparsity structure $\F$ and (ii) the weights $\w_{m,k}$
mapping the latent variables into the observations are correct. Since
the methods provide solutions of a varying number of factors and do
not preserve the order of factors, we use a similarity measure
\citep{Knowles11} that chooses an optimal re-ordering and sign for the
factors found by the models, and then measures the mean-square error.

\begin{figure}[t]
\centering
\begin{tabular}{c}
\includegraphics[scale=0.35]{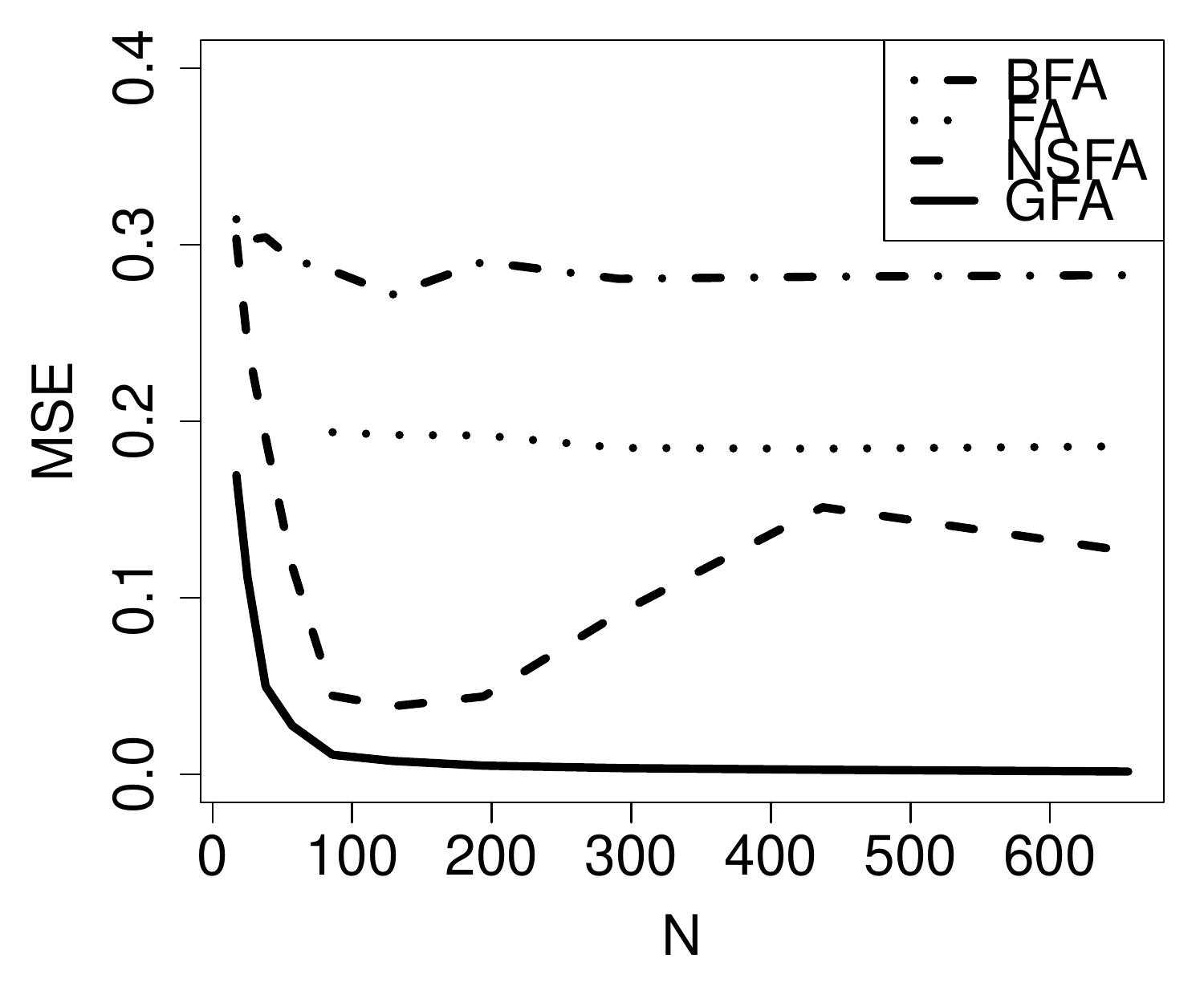} \\
\includegraphics[scale=0.35]{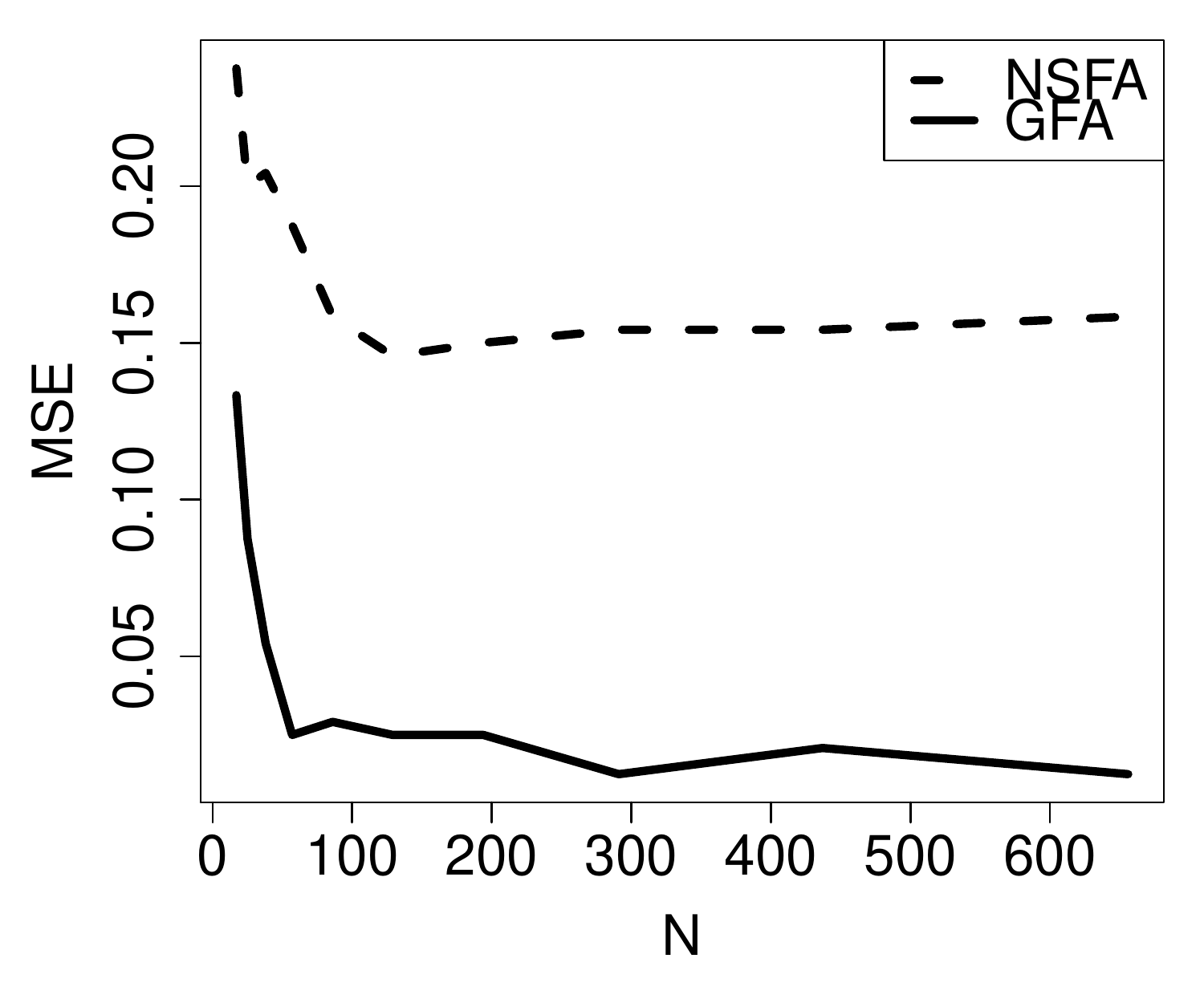}
\end{tabular}
\caption{{\bf Top:} Difference (mean square error MSE) between the
  estimated and true loading matrix $\W$ as a function of the
  sample size N. Our Bayesian group factor analyzer (GFA) directly
  solving the GFA problem is consistently the best, beating
  state-of-the-art nonparametric sparse factor analysis (NSFA), regular
  factor analysis (FA) and Bayesian FA (BFA). It is
  worthwhile to notice that increasing the sample size does not help
  the alternative methods to obtain the correct solution.  {\bf
    Bottom:} Difference between the estimated and true factor
  activity matrix $\F$, shown for the two methods providing sparse
  representations. Again GFA outperforms NSFA. 
}
\label{fig:toydata1}
\end{figure}

We start by measuring property (ii), by inspecting the similarity of
the true and learned loading matrix $\W$. The GFA finds the mappings much
more accurately than the alternative methods
(Fig.~\ref{fig:toydata1}). Next we inspect the property (i), the
similarity of the true and estimated $\F$, again using the same measure
but for binary matrices. Since FA and BFA do not enforce any kind of
sparsity within factors, we only compare GFA and NSFA. For GFA we
obtain $\F$ by thresholding the ARD weights using
\eqref{eq:threshold}, whereas for NSFA we set $\f_{m,k}=1$ if any
weight within $\w_{m,k}$ is non-zero. GFA again outperforms NSFA which
has not been designed for the task. By adaptive thresholding
of the weight sets of NSFA it would be possible to reach almost the
accuracy of GFA in discovering $\F$, but as shown by the comparison of
$\W$ the actual factors would still be incorrect.

Next we proceed to inspect how well GFA can extract the correct
structure from different kinds of data collections, with a particular
focus on discovering whether it biases specific types of factors. If
no such bias is found, the experiments give empirical support that the
method solves the GFA problem in the right way. For this purpose we
constructed data sets with a fixed number of samples ($N=100$) and a
varying number of views $M$ that are all $D=10$ dimensional. We
created data collections with three alternative distributions over the
different kinds of factors, simulating possible alternatives
encountered in real applications.  The first distribution type has one
factor of each possible cardinality and the second shows a
power-law distribution with few factors active in many views and
many very sparse factors. Finally, the third type has a uniform
distribution over the subsets, resulting in the cardinality
distribution following binomial coefficients.

\begin{figure*}[t]
 \centering
\begin{tabular}{cccc}
(a) & (b) & (c) & (d) \\
\includegraphics[scale=0.22]{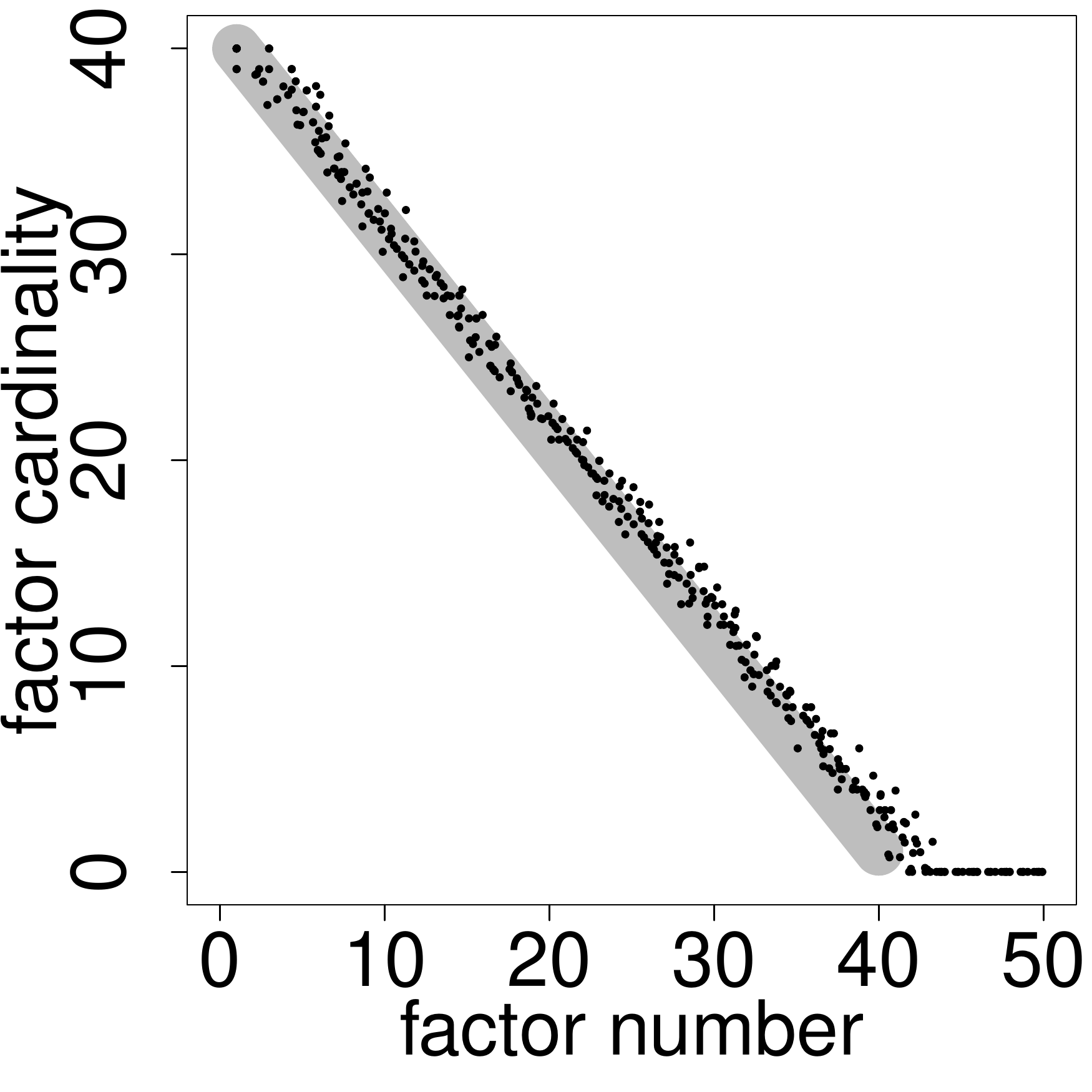} &
\includegraphics[scale=0.22]{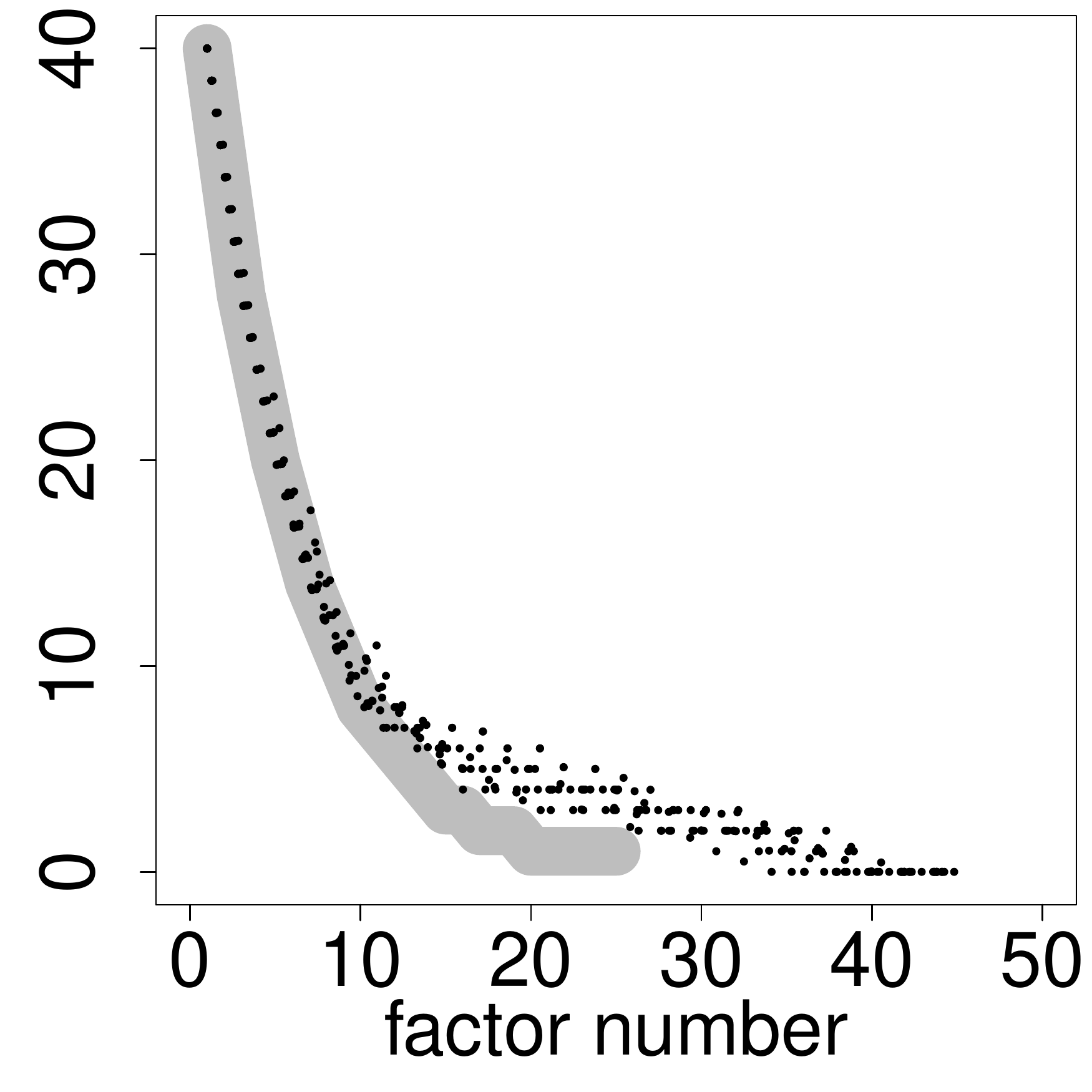} &
\includegraphics[scale=0.22]{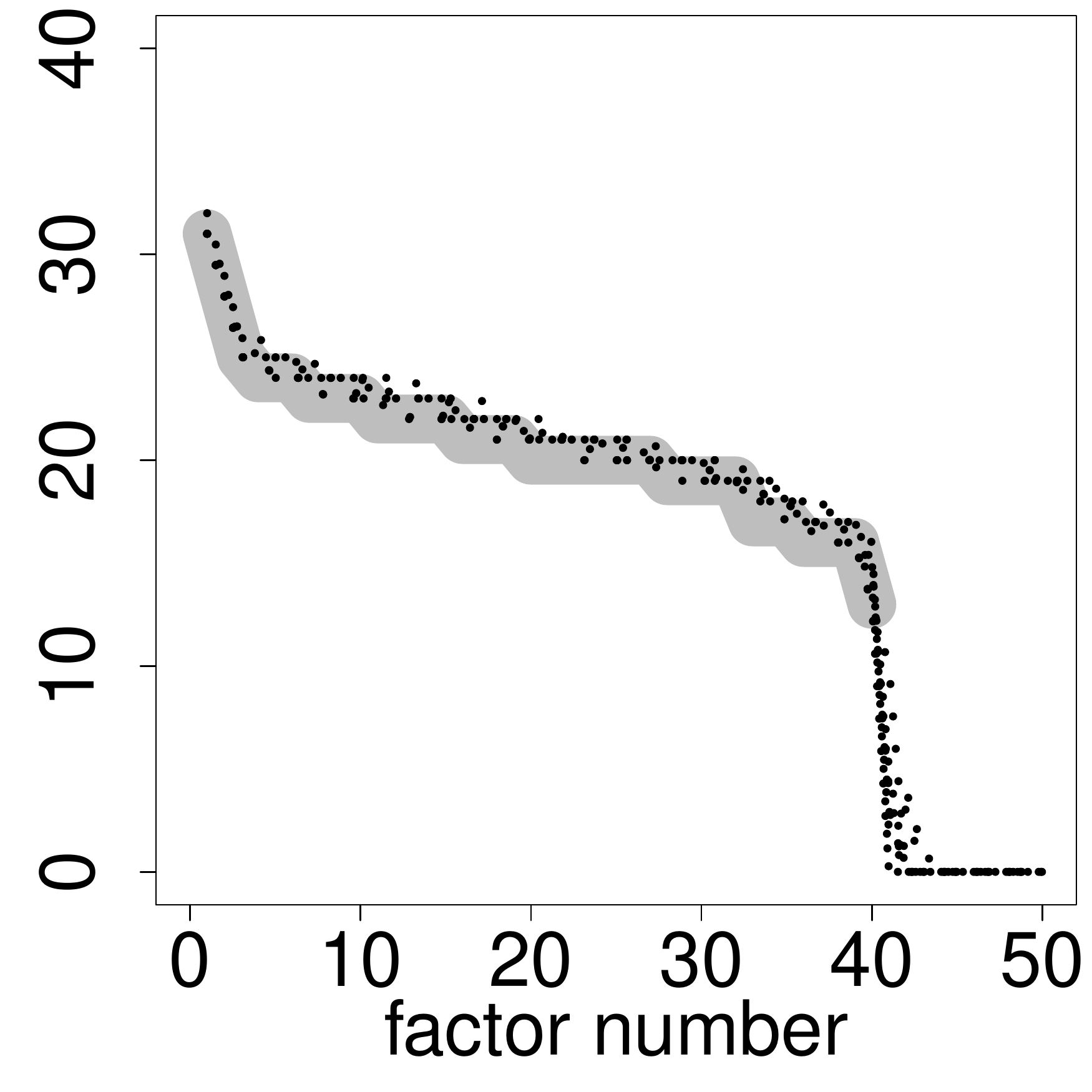} &
\includegraphics[scale=0.22]{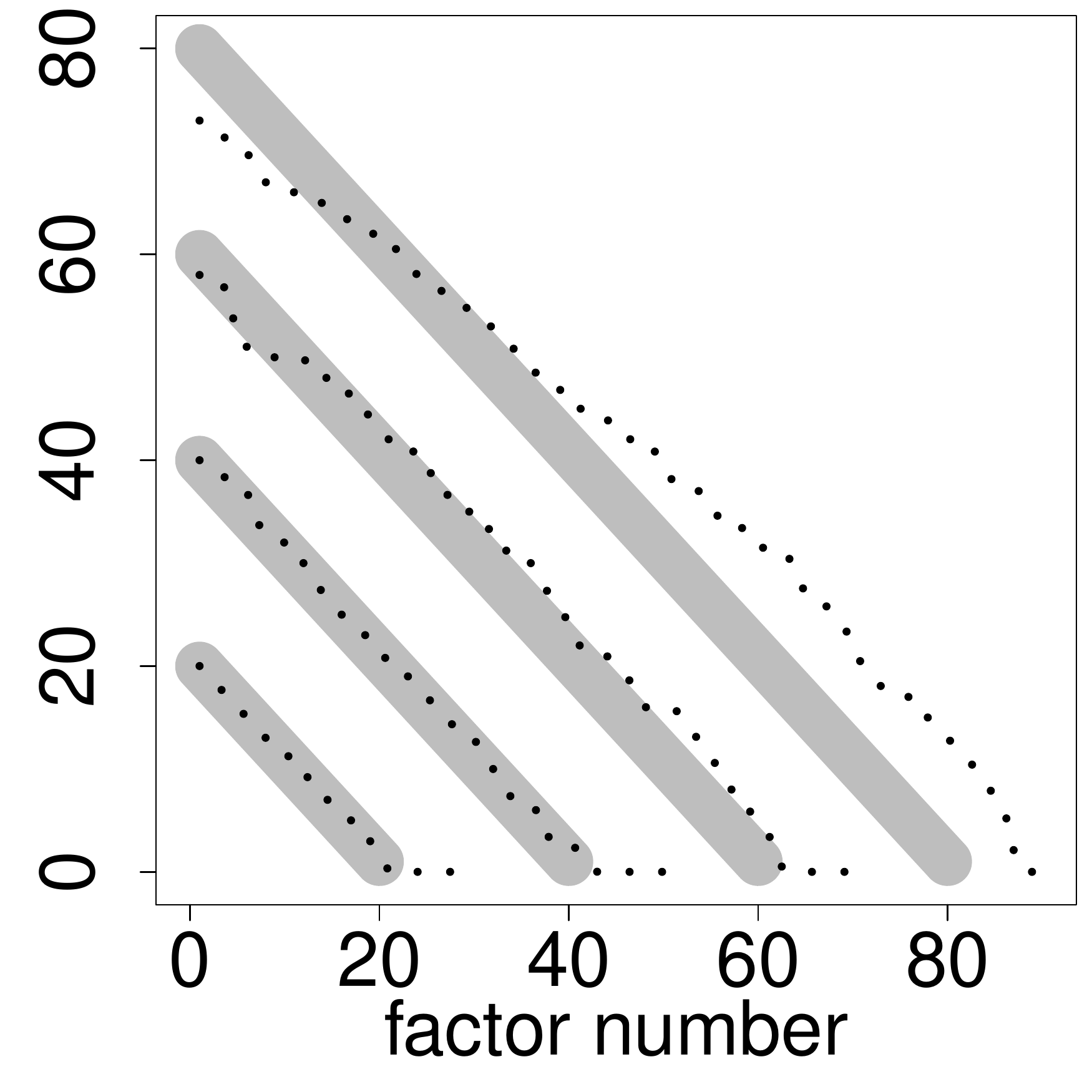} 
\end{tabular}
\caption{The GFA model finds the correct sparsity structure $\F$ for
  three very different distributions over different types of factors
  in the data. The thick grey line shows the true latent structure as
  the cardinalities (number of active views) of the factors in the
  decreasing order, and the overlaid curves show results of 10 runs on
  different simulated data. The first three plots show results for
  $M=40$ views with three different distributions of factor types
  ((a): uniform over the cardinality, (b): power law
  over the cardinality, (c): uniform over the view
  combinations), and for all cases the model learns the correct
  structure. The last plot (d) shows that the behavior is
  consistent over a wide range of $M$ (the different curves) and only
  starts to break down when $M$ approaches the sample size $N=100$.}
\label{fig:toy2}
\end{figure*}

Figure~\ref{fig:toy2} shows the true and estimated distribution of
factor cardinalities (the number of active views in a factor) for the
different data collections. 
For all three cases, the model finds the correct
structure for $M=40$ views, and in particular models correctly both
types of factors: those shared by several views and those specific to only one or a
few. Besides checking for the correct cardinalities, we inspected
that the actual factors found by the model match the true ones.
We then proceeded to demonstrate (Fig.~\ref{fig:toy2} (d)), for the case with uniform
distribution over factor cardinalities, that the finding holds for all
numbers of views below $M=60$ for this case with just $N=100$ samples;
for other distributions the results are similar (not shown).

\section{Application scenarios and interpretation}

Next, we apply the method to brain activity and drug response
analysis, representative of potential use cases for GFA, and show how
the results of the model can be interpreted. Both applications contain
data of multiple views (7 and 4, respectively) and could not be
directly analyzed with traditional methods. The number of views is
well within the range for which the model was demonstrated above to
find the correct structure from the simulated data.

Both applications follow the same analysis procedure, which can be
taken as our practical guidelines for data-analysis with GFA.  First, the
model is learned with sufficiently many factors, recognized as a
solution where more than a few factors are left at zero. Solutions
where all the factors are in use cannot be relied upon, since then it
is possible that a found factor describes several of the underlying
true factors. After that, the factor activity matrix $\F$, ordered
suitably, is inspected in search for interesting broad-scale
properties of the data. In particular, the number of factors specific
to individual views is indicative of the complexity of residual
variation in the data, and factors sharing specific subsets of views
can immediately reveal interesting structure.  Finally, individual
factors are selected for closer inspection by ordering the factors
according to an interest measure specific to the application. We
demonstrate that measures based on both $\Z$ and $\W$ are meaningful.

\subsection{Multi-set analysis of brain activity}

We analyze a data collection where the subject has been exposed to
audiovisual music stimulation. The setup is uniquely multi-view; each
subject experienced the same three pieces of music seven times as
different variations. For example, in one condition the subjects viewed
a video recording of someone playing the piece on piano, in one
condition they only heard a duet of singing and piano, and in one they
saw and heard a person reading out the lyrics of the song.

The $M=7$ different exposure conditions or stimuli types were used as
views, and we sought factors tying together the different conditions
by applying GFA to a data set where the fMRI recordings of the 10
subjects were concatenated in the time direction. Each sample
($N=1162$) contained, for each view, activities of $D_m=32$ regions of
interest (ROI) extracted as averages of local brain regions. The set
of $32$ regions was chosen by first picking the five most correlating ROIs for
each of the $21$ possible pairs of listening conditions and then taking
the union of these choices.

\begin{figure}[t]
\centering
\begin{tabular}{cc}
\includegraphics[scale=0.38]{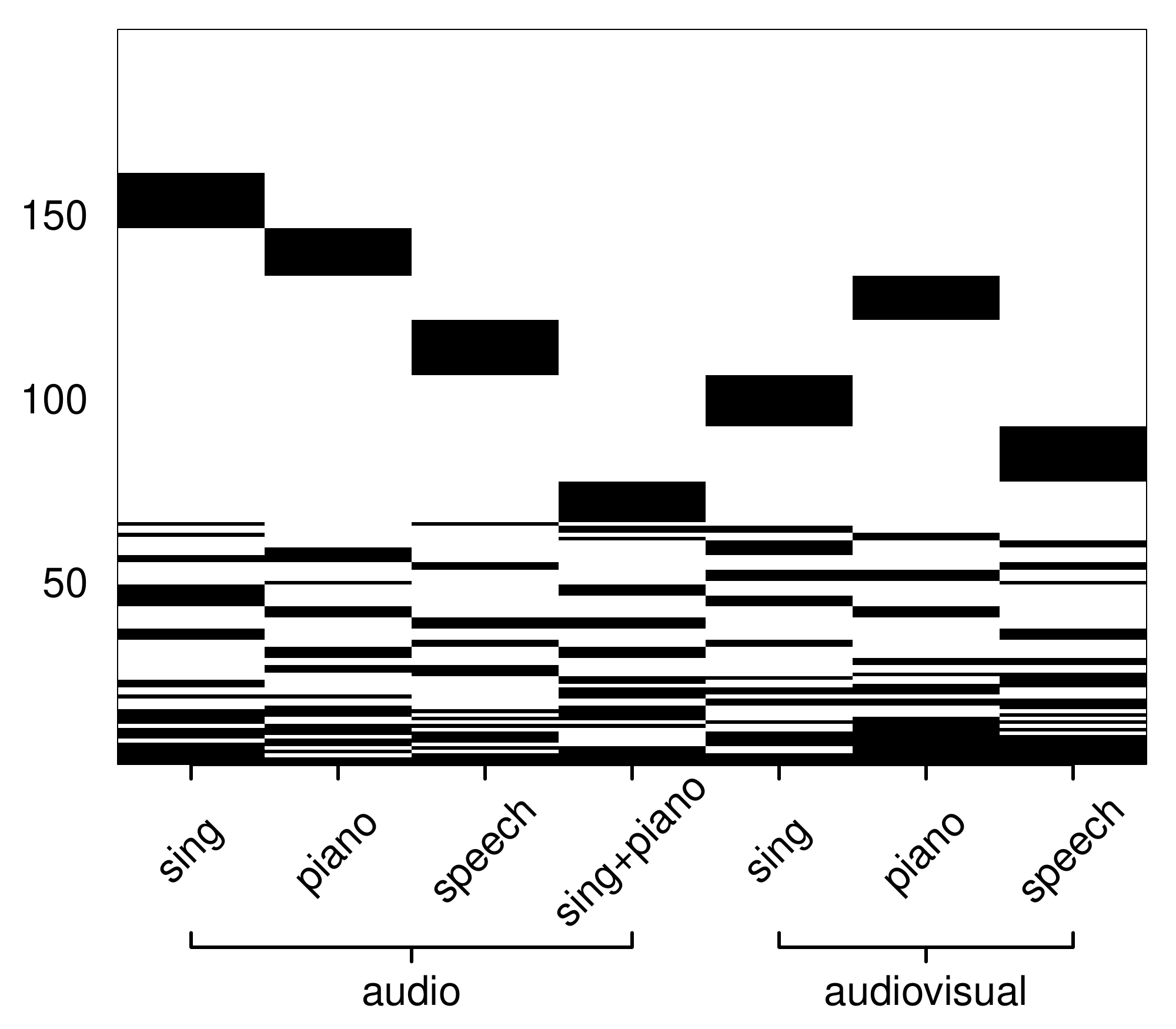} \\
\includegraphics[scale=0.38]{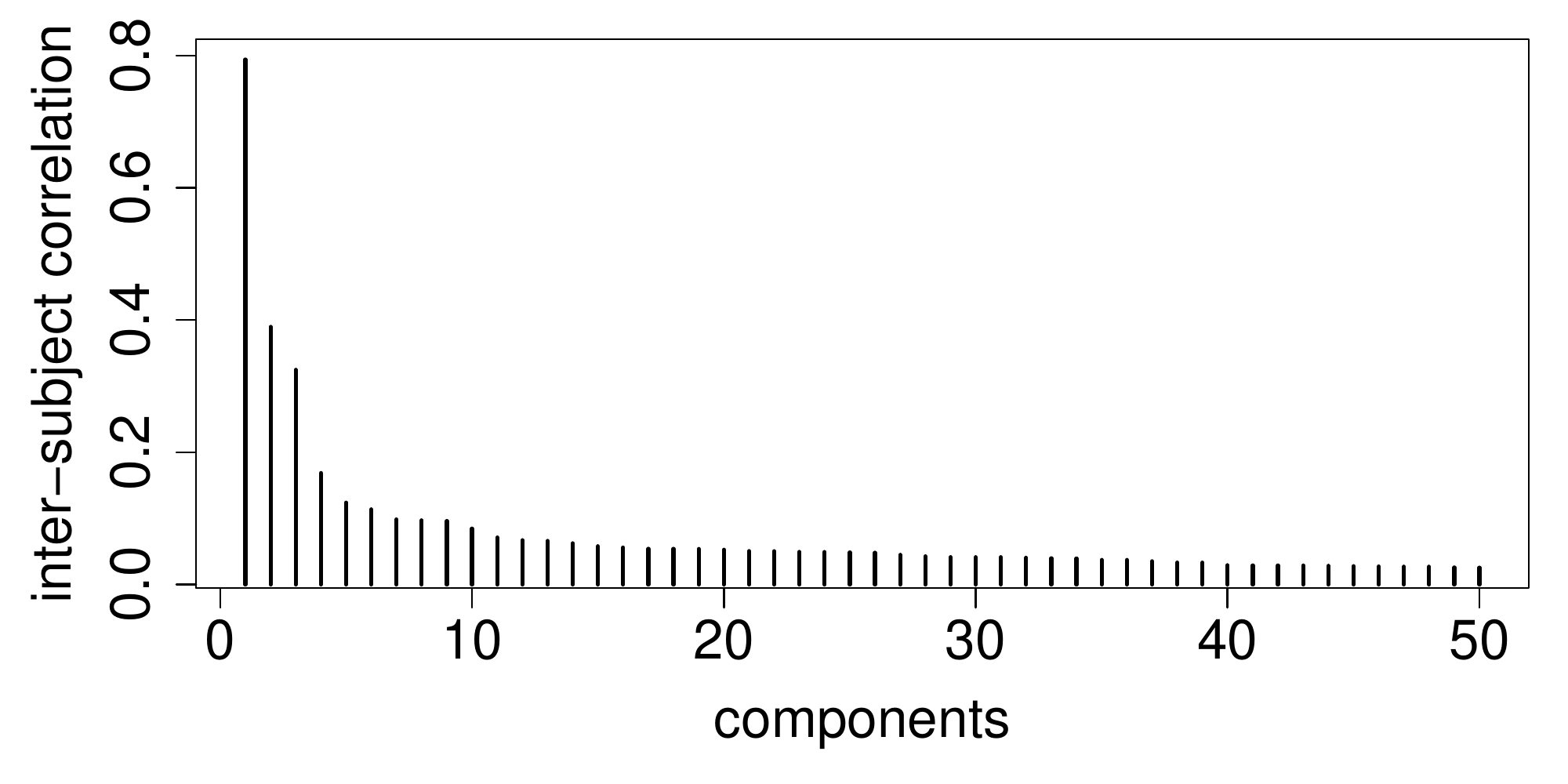}
\end{tabular}
\caption{Multi-set analysis of brain activity. {\bf Top:} The matrix $\F$ of factors (rows) across views (columns),
  indicating the dependencies between the views. The different views
  correspond to different versions of the same songs being played for
  the subject. {\bf Bottom:} Sorting the factors as a function of
  inter-subject correlation reveals which factors robustly capture the
  response to the stimulus.}
\label{fig:brainF}
\end{figure}

The factor activity matrix $\F$ of factors times views, of a 200-component
solution, is shown in Figure~\ref{fig:brainF}. The number of factors is
sufficient, as demonstrated by the roughly 40 empty factors, and we
see that the data contain large blocks of view-specific factors
suggesting a lot of noise in the data.  To more closely examine the
factors, we chose to study factors that are coherent over the
users. We split the latent variables $\z_k \in \mathbb R^N$
corresponding to each factor into 10 sequences corresponding to the
samples of the 10 subjects, and measured the inter-subject correlation
(ISC) of each factor. The factors were then ranked according to ISC
(Fig.~\ref{fig:brainF}, bottom), revealing a few components having
very high synchrony despite the model being ignorant that the data
consisted of several users.

The strongest ISC correlation is for a component shared by all views.
It captures the main progression of the music pieces irrespective of
the view. A closer inspection of the weight vectors reveals that the
responses in the different views are in different brain regions
according to the modality; the four conditions with purely auditory
stimuli have weights revealing auditory response, whereas the three
conditions with also visual stimulation activate also vision-related
regions. The second-strongest ISC correlation is for a component
shared by just two views, speech under both audiovisual and purely
auditory conditions. That is, the component reveals a response to
hearing recitation of the song lyrics instead of actual music as in
the other conditions.

The solution is consistent with how GFA is meant to work. In this
specific application, it might be useful to additionally apply a model
where solutions with similar $\w_{m,k}$ across all $m$ with
$\f_{m,k}=1$ would be favored. The components would then be directly
interpretable in terms of specific brain
activity patterns, in addition to the time courses.

\subsection{Multi-set analysis of drug responses}

The second case study is a novel chemical systems biology application,
where the observations are drugs and the first $M-1$ views are responses
of different cell lines (each being a proxy to a different disease) to
the specific drug. The $M$th view contains features describing
chemical properties of the drug, derived from its structure. The
interesting questions are can GFA find factors relating drug structures
and diseases, or relating different cell lines which are here different cancer
subtypes.

As the biological views, we used activation profiles over
$D_1=D_2=D_3=1321$ gene sets in $M-1=3$ cancer cell lines compared to
controls, from \citep{Lamb06}. The $4$th (chemical) view consisted of
a standard type of descriptors of the 3D structure of the drugs,
called VolSurf ($D_4=76$). The number of drugs for which observations
had been recorded for all cell lines was $N=684$.

\begin{figure}[t]
\centering
\begin{tabular}{cc}
\includegraphics[scale=0.35]{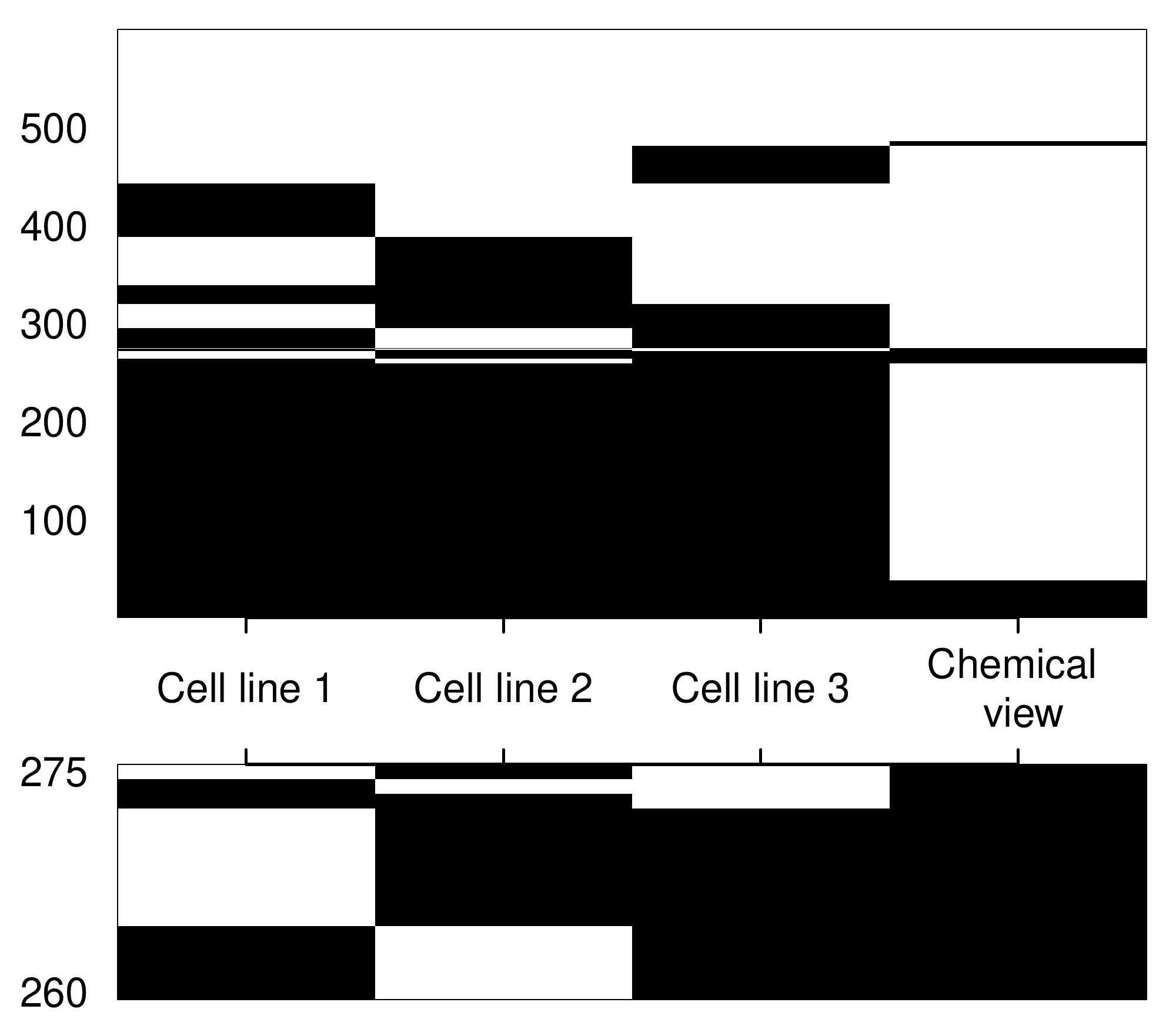} \\
\includegraphics[scale=0.29]{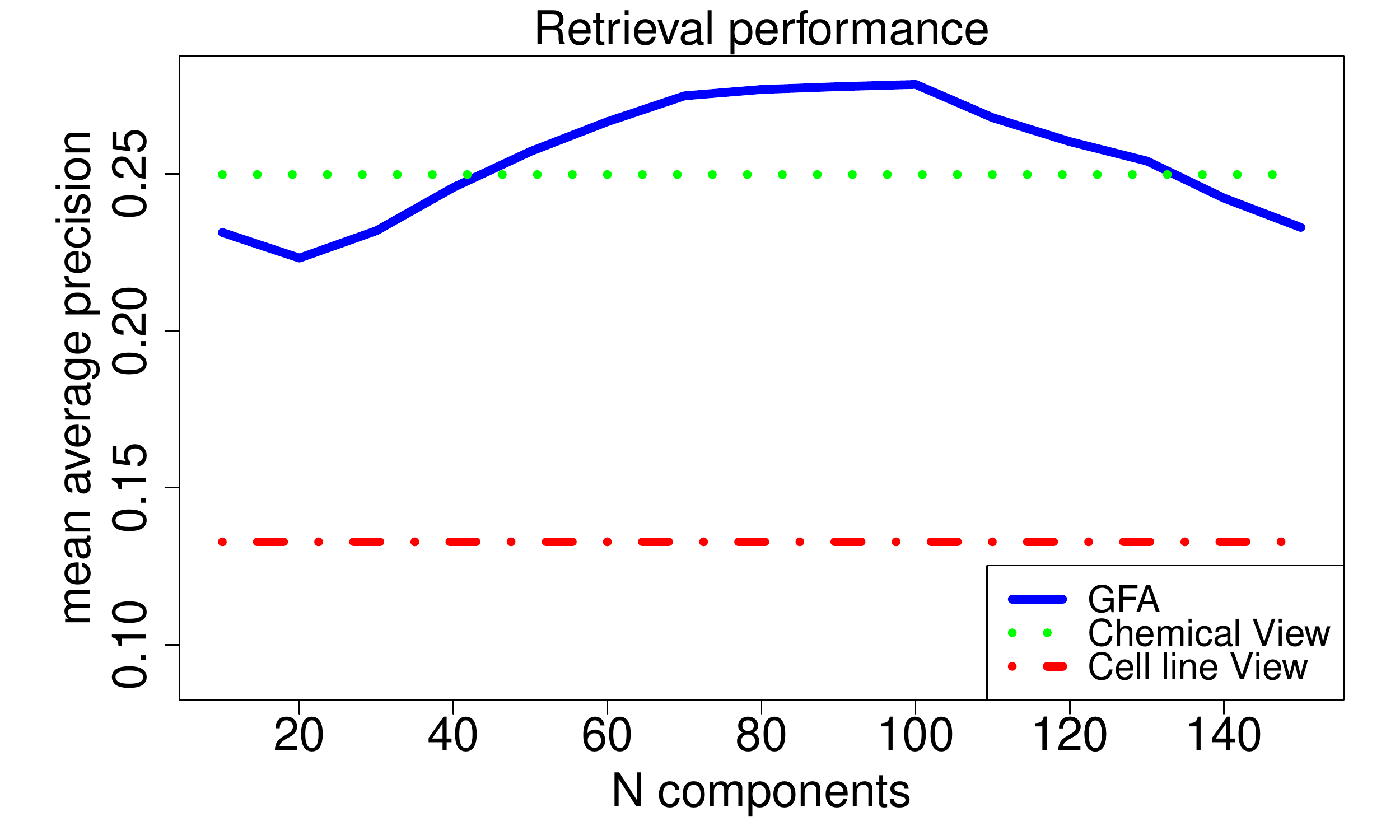} 
\end{tabular}
\caption{Multi-set analysis of drug responses. \textbf{Top:} Factor activity matrix of the factors (rows) against
  the 3 biological views (columns), cell lines HL60, MCF7, PC3, and the chemical
  view of drug descriptors. The small matrix at the bottom shows a
  zoomed inset to an interesting subset of the factors. \textbf{Bottom:} Mean average precision of
  retrieving drugs having similar effects (targets), based on the first N GFA factors. Integration of the
  data sources gives a significantly higher retrieval performance than any data
  source separately.}
\label{fig:chembioF}
\end{figure}

The factor-view matrix $\F$ of a 600-component GFA solution is
shown in Figure~\ref{fig:chembioF} (top). The number of components is
large enough since there are over 100 empty factors.  Four main types
of factors were found: (i) Factors shared by the chemical view and one
(or two) cell lines (zoomed inset in Fig~\ref{fig:chembioF},
top). They give hypotheses for the specific cancer variants. (ii)
Factors shared by all cell lines and the chemical space, representing
effects specific to all subtypes of cancer.
(iii) Factors shared by all cell lines but not the
chemical space. They are drug effects not captured by the specific
chemical descriptors used. The fact that there is a large block of
over 200 of these factors fits well with the known fact that VolSurf features are
a very limited description of the complexity of drugs.  (iv) Factors
specific to one biological view. These represent either ``biological
noise'' or then drug effects specific to that cancer subtype, again
not captured by the VolSurf features. Finally, the small set of
components active only in the chemical view correspond to structure
in VolSurf having no drug effect.

We inspected some of the factors more carefully, and more detailed
biological analysis is on-going. The first factor of type (i), shared
by one cell line and the chemical descriptors, activates genes linked
to inflammatory processes on the biological side, and is active in
particular for non-steroidal anti-inflammatory drugs (NSAIDs),
especially ibuprofen-like compounds, which are known to have
anti-cancer effects \citep{Ruegg03}. Of the factors shared by all cell lines (type
iii), the one with the highest norm of the weight vectors shows strong
toxic effects on all cell lines, being linked to stopping of cell growth
and apoptosis. In summary, these first findings are well in line with
established knowledge, and digging into the further components is
on-going.

We next validated quantitatively the ability of the model to discover
biologically meaningful factors. We evaluated the performance of the
found set of factors in representing drugs in the task of retrieving
drugs known to have similar effects (having the same
targets). 

We represented each drug with the corresponding vector in the latent
variable matrix $\Z$, and used correlation along the vectors as a
similarity measure when retrieving the drugs most similar to a query
drug. Retrieval performance was measured as the mean average precision
of retrieving drugs having similar effects (having the same targets).
As baselines we computed the distances in only the biological views or
in only the chemical view. Representation by the GFA factors
outperforms significantly (t-test, $p<0.05$) using either space separately
(Fig.~\ref{fig:chembioF}, bottom). The experiment was completely data
driven except for one bit of prior knowledge: As the chemical space is
considered to be the most informative about drug similarity, the factors
were pre-sorted by decreasing Euclidean norm of the weight vectors $\w_k$
in the chemical space.

\section{Discussion}

We introduced a novel problem formulation of finding
factors describing dependencies between data sets or views, extending
classical factor analysis which does the same for variables. The task
is to provide a set of factors explaining dependencies between all
possible subsets of the views. For solving the problem, coined group
factor analysis (GFA), we provided a group-wise sparse Bayesian factor
analysis model by extending a recent CCA formulation by
\citet{Virtanen11icml} to multiple views. The model was demonstrated to
be able to find factors of different types, including those specific to just one
view and those shared by all views, equally well even for high numbers
of views. We applied the model to data analysis in new kinds of
application setups in neuroimaging and chemical systems biology.

The variational approximation used for solving the problem is
computationally reasonably efficient and is directly applicable to
data sets of thousands of samples and several high-dimensional views,
with the main computational cost coming from a high number of factors
slowing down the search for an optimal rotation of the factors. It
would be fruitful to develop (approximative) analytical solutions for
optimizing Eq.~\ref{eq:rotation} necessary for the model to converge
to the correct sparsity structure, which would speed up the algorithm
to the level of standard Bayesian PCA/FA.

The primary challenge in solving the GFA problem is in correctly
detecting the sparsity structure. Our solution was demonstrated to be
very accurate at least for simulated data, but it would
be fruitful to study how well the method fares in comparison with
alternative modeling frameworks that could be adapted to solve the GFA
problem, such as the structured sparse matrix factorization 
by \citet{Jia10} or extensions of nonparametric sparse factor analysis
\citep{Knowles11} modified to support group sparsity.  It could also be
useful to consider models that are group-wise sparse but allow
sparsity also within the active factor-view groups or sparse
deviations from zero for the inactive ones, with model structures
along the lines \citet{Jalali10} proposed for multi-task
learning.

\subsubsection*{Acknowledgments}

We acknowledge the Academy of Finland (project numbers 133818, 140057 and
the AIRC Center of Excellence), the aivoAALTO project, FICS, and PASCAL2
EU Network of Excellence for funding.

We are deeply grateful to Dr. Krister Wennerberg (FIMM at
University of Helsinki) for helping us in interpretation of discovered
factors for the drug structure-response application. We also thank
Dr. Juha Salmitaival, Prof. Mikko Sams, and MSc. Enrico Glerean
(BECS at Aalto University) for providing the data used in the
neuroscientific experiment.

\end{document}